\documentclass{article}

\usepackage{microtype}
\usepackage{graphicx}
\usepackage{subfigure}
\usepackage{booktabs}

\usepackage{hyperref}

\usepackage{times}
\usepackage{latexsym}
\usepackage{enumitem}

\usepackage[accepted]{icml2025}

\usepackage{amsmath}
\usepackage{amssymb}
\usepackage{mathtools}
\usepackage{amsthm}

\usepackage[capitalize,noabbrev]{cleveref}

\theoremstyle{plain}

\theoremstyle{definition}

\theoremstyle{remark}

\usepackage{inconsolata}
\usepackage[utf8]{inputenc}
\usepackage[T1]{fontenc}
\usepackage{hyperref}
\usepackage{url}
\usepackage{booktabs}
\usepackage{amsfonts}
\usepackage{nicefrac}
\usepackage{microtype}
\usepackage{xcolor}
\usepackage{graphicx}
\usepackage{booktabs}
\usepackage{array}
\usepackage{multicol}
\usepackage{multirow}
\usepackage{amssymb}
\usepackage{pifont}
\usepackage{tikz}
\usepackage{xcolor}
\usepackage{xspace}
\usepackage{amsmath}

\usepackage[textsize=tiny]{todonotes}

\newcommand{\method}{\textsc{LC-R1}\xspace}
\newcommand{\extractor}{LC-Extractor\xspace}

\usepackage{fontawesome5}
\usepackage{xcolor}

\usepackage[breakable,skins]{tcolorbox}
\usepackage{booktabs}
\usepackage{float}
\usepackage{cuted}
\usepackage{caption}

\definecolor{boxbackground}{RGB}{255,255,255}
\definecolor{boxborder}{RGB}{200,200,200}
\definecolor{accentblue}{RGB}{0,114,187}
\usepackage{algorithm}
\usepackage{amsmath}
\usepackage{lipsum}
\usepackage{algorithm}
\usepackage{algorithmic}
\usepackage{amsfonts}
\usepackage{amssymb}
\usepackage{subcaption}
\newtcolorbox{promptbox}[2][]{
    enhanced,
    breakable,
    boxsep=5pt,
    left=9pt,
    right=7pt,
    top=5pt,
    bottom=5pt,
    colback=boxbackground,
    colframe=boxborder,
    boxrule=0.5pt,
    arc=4pt,
    frame hidden,
    borderline west={3pt}{0pt}{accentblue},
    shadow={0.5pt}{0.5pt}{1.5pt}{black!10},
    fontupper=\normalsize,

    title=#2,
    colbacktitle=accentblue,
    coltitle=white,
    fonttitle={\fontsize{9}{11}\selectfont\bfseries},

    attach boxed title to top left={yshift=-2.5mm, xshift=3.2mm},
    boxed title style={
        enhanced,
        left=3pt,
        right=3pt,
        top=1pt,
        bottom=1pt,
        boxsep=2pt,
        arc=3pt,
        boxrule=0pt,
        colback=accentblue,
    },
    #1
}

\usepackage{fancyhdr}
\icmltitlerunning{Optimizing Length Compression in Large Reasoning Models}

\begin{document}

\twocolumn[
\icmltitle{ Optimizing Length Compression in Large Reasoning Models}

\icmlsetsymbol{equal}{*}

\begin{icmlauthorlist}
\icmlauthor{Zhengxiang Cheng}{}
\icmlauthor{Dongping Chen}{umd}
\icmlauthor{Mingyang Fu}{}
\icmlauthor{Tianyi Zhou}{}

\end{icmlauthorlist}

\icmlaffiliation{umd}{University of Maryland}

\icmlcorrespondingauthor{Tianyi Zhou}{tianyi.david.zhou@gmail.com}

\icmlkeywords{Machine Learning, ICML}

\vskip 0.3in
]
\printAffiliationsAndNotice{}

\begin{abstract}

Large Reasoning Models (LRMs) have achieved remarkable success, yet they often suffer from producing unnecessary and verbose reasoning chains. 
We identify a core aspect of this issue as \textit{``invalid thinking''}--- models tend to repeatedly double-check their work after having derived the correct answer. 
To address this specific inefficiency, we move beyond the general principles of Efficacy and Efficiency to propose two new, fine-grained principles: Brevity, which advocates for eliminating redundancy, and Sufficiency, which ensures critical reasoning steps are preserved.
Guided by these principles, we introduce \method, a post-training method based on Group Relative Policy Optimization (GRPO).
\method employs a novel combination of a \textit{Length Reward} for overall conciseness and a \textit{Compress Reward} that is specifically designed to remove the invalid portion of the thinking process. 
Extensive experiments on multiple reasoning benchmarks demonstrate that \method achieves a significant reduction in sequence length (\textasciitilde 50\%) with only a marginal (\textasciitilde 2\%) drop in accuracy, achieving a favorable trade-off point on the \textit{Pareto frontier} that prioritizes high compression. Our analysis further validates the robustness of \method and provides valuable insights for developing more powerful yet computationally efficient LRMs.
Our code is released at https://github.com/zxiangx/LC-R1.

\end{abstract}

\section{Introduction}

\begin{figure}[!t]
    \centering
    \includegraphics[width=1\linewidth]{Figures/FirstPage_big.pdf}
    \caption{Comparison between inefficient reasoning model and efficient model. The former tends to make a verbose self-check process after having derived the correct answer corresponding to the given question. The model trained with \method get more efficient reasoning process to get correct answer, without any invalid thinking process.}
    \label{fig:simple-case}

\end{figure}

 \begin{figure*}[!t]
    \centering
    \includegraphics[width=1\linewidth]{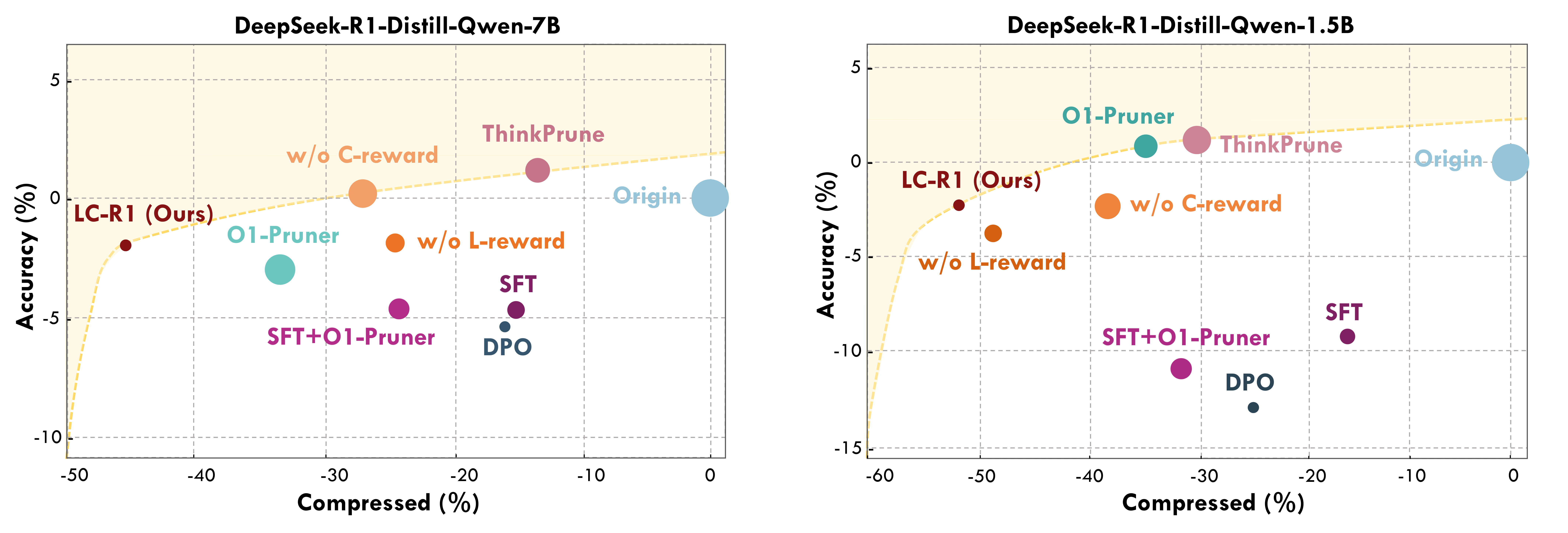}
    \vspace{-1em}
    \caption{Pareto analysis of the Efficacy-Efficiency trade-off of different methods on two reasoning models. The x-axis represents the reasoning length change, and the y-axis shows the accuracy change, relative to the original model (defined in Eq. \ref{metric}), with the top-left corner representing the ideal position. A smaller and darker marker indicates a higher Valid Thinking (VT) rate (defined in Eq. \ref{valid thinking}), signifying a more efficient thinking process. Compared to other methods also on the pareto frontier, \method achieves a more favorable trade-off, attaining a substantially higher compression rate at the cost of a minimal drop in accuracy, and it also achieves a higher VT rate. The sub-optimal performance of our ablation variants (w/o C-reward, w/o L-reward) further proves the criticality of our dual-reward designs. }
    \label{fig:pareto}
\end{figure*}

Recent \textit{``long-thought''} Large Reasoning Models (LRMs), such as OpenAI's O1 \citep{jaech2024openai} and Deepseek-R1 \citep{deepseekai2025deepseek-r1}, represent a significant paradigm extension of foundational Chain-of-Thought (CoT) techniques \citep{wei2023chain-of-thought}. Fine-tuned with Reinforcement Learning (RL), these models iteratively refine solutions to achieve unprecedented performance in complex reasoning tasks like mathematics and programming \citep{sun2025olym-math, gu2024cruxeval}.
However, with the improvement of \textit{``deep thinking''} ability, a prominent problem is the excessive consumption of computing resources during the reasoning process \citep{chen2025reasoning-era-survey-long,aggarwal2025l1controllinglongreasoning}. Specifically, existing models tend to generate lengthy and even unnecessary chains of reasoning when solving problems with low complexity or clear solution paths. This phenomenon, termed ``\textit{overthinking}'', manifests as models consuming far more computational resources than the problem itself requires to reach the correct conclusion \citep{chen2024overthinking,sui2025stop-overthinking-survey,cuadron2025overthinking-danger}. Therefore, one critical problem arises:

\begin{center}
    \textbf{\textit{How can we maintain high reasoning efficacy while significantly improving efficiency?}}
\end{center}

Prior works have approached this by fine-tuning on shorter demonstrations (SFT) \citep{chen2024overthinking}, constructing preference datasets for conciseness \citep{luo2025adar1,shen2025dast-difficultyadaptiveslowthinkinglarge}, or integrating length-penalties into RL \citep{hou2025thinkprune,luo2025o1prunerlengthharmonizingfinetuningo1like,kimi}. However, these methods often treat the reasoning process as a black box, penalizing length without analyzing the internal structure of the thoughts themselves.

To address this gap, we delve into the structure of ``\textit{overthinking}'' and identify a specific pattern: models frequently engage in redundant \textit{``double-checking''} after having already derived the correct answer. We term this phenomenon ``\textit{invalid thinking}'', as shown in Figure \ref{fig:simple-case}. To quantify it, we introduce a new metric, \textbf{Valid Thinking (VT) Rate}, which measures the proportion of the reasoning process that is essential for reaching the initial correct conclusion.

Guided by this insight, we propose two fine-grained principles: \textbf{Brevity} (eliminating redundancy) and \textbf{Sufficiency} (preserving necessary steps). We then introduce \method, a GRPO-based post-training method that operationalizes these principles. \method uniquely combines a \textbf{Length Reward} for overall conciseness with a novel \textbf{Compress Reward} designed to directly guide the model to terminate the thinking process upon deriving the correct answer.

We conduct comprehensive experiments on two reasoning models across seven benchmarks. Empirical results show that \method achieves a more favorable trade-off between efficacy and efficiency than prior methods as shown in Figure \ref{fig:pareto}. Specifically, with only a 2\% drop in accuracy, our method attains a 50\% reduction in sequence length on average. Ablation study also demonstrates the indispensability of both Length Reward and Compress Reward for achieving efficient reasoning. Further study shows that our method achieves efficient compression without impairing the exploration ability of model, and the efficiency can generalize to various difficulty problems. In conclusion, our contribution can be summarized as follows:

\begin{table*}[!t]
 \newcolumntype{C}[1]{>{\centering\arraybackslash}p{#1}}
    \centering
    \caption{Valid Thinking Rate of current \emph{state-of-the-art} Large Reasoning Models.  Nemotron indicates Llama-3.3-Nemotron-Super-49b-v1. Results manifest a low VT rate on all these models, highlighting the phenomenon of ``\textit{invalid thinking}''.}
    \label{tab:verbosity_pilot_study}
    \scalebox{1}{
    \begin{tabular}{l|C{1.2cm}|*{5}{C{1.4cm}}}
     \toprule[1.5pt]
    \textbf{Model} & \textbf{Avg.} & \textbf{AIME25} & \textbf{AMC} & \textbf{GSM8K} & \textbf{MATH500} & \textbf{Olympiad} \\
    \midrule
    \textbf{Qwen-3-32B} & \textbf{57.5} & 73.8 & 58.8 & \textbf{53.8} & \textbf{46.6} & \textbf{51.5} \\
    \textbf{QwQ-32B} & 59.2 & 70.8 & \textbf{58.2} & 54.1 & 53.1 & 59.6 \\
    \textbf{DeepSeek-R1} & 65.3 & 66.5 & 71.8 & 64.2 & 59.8 & 64.0 \\
    \textbf{Nemotron} & 60.8 & \textbf{62.1} & 64.1 & 63.1  & 56.6 & 58.1 \\
    \bottomrule[1.5pt]
    \end{tabular}}
\end{table*}

\begin{itemize}
    \item We analyze the thinking process of current competitive reasoning model and find the phenomenon of ``\textit{invalid thinking}'' : It takes a large portion of thinking process to double check after having derived the correct answer, making the reasoning verbose and inefficient.
    \item We propose two novel principles: \textbf{Brevity} and \textbf{Sufficiency}, and design a GRPO-based method \textbf{\method} for LRM post-training to strike a balance between Brevity and Sufficiency, pruning invalid thinking while compressing overall sequences at the same time.
    \item Through comprehensive experiments, we validate the effectiveness of \method to get a better trade-off between Efficacy and Efficiency, and conduct further analyses on the deep impact of compression, proving the robustness of \method to various difficulties and providing insights for future works.
\end{itemize}

\section{Preliminary: Compression and Efficienct Reasoning Models}
\subsection{Motivation: Quantifying Redundant Reasoning}

A common paradigm for Large Reasoning Models (LRMs) involves a \textbf{thinking process} (\emph{i.e.}, step-by-step rationale) that precedes the final answer. While effective for accuracy, we observe a consistent inefficiency: models often derive the correct answer early in their thinking process but continue with lengthy and redundant verification steps. We term this subsequent, non-essential reasoning \textit{``Redundant Sequence''}.

To formalize this, we define the \textit{Valid Thinking (VT)} rate, a metric focusing on the model's thinking process:
\begin{equation}
\label{valid thinking}
    \text{VT}=\frac{|\text{Tokens in Valid Thinking}|}{|\text{Total tokens in Thinking Process}|}
\end{equation}

where \textit{``Valid Thinking''} comprises the tokens from the start of the thinking process until the correct answer is first derived. To automate this measurement, we utilize a lightweight parser, \extractor, whose implementation details are provided in Section \ref{experiment}.

we evaluated four \textit{state-of-the-art} LRMs---Qwen3-32b \citep{qwen3}, QwQ-32b \citep{qwq32b}, Deepseek-R1 \citep{deepseekai2025deepseek-r1}, and Llama-3.3-nemotron-super-49b-v1 \citep{bercovich2025llama-nemotron}---across five math benchmarks: AIME25, MATH500, GSM8K, AMC, OlympiadBench.
Our analysis reveals a universal and severe \textit{overthinking} problem. As shown in Table \ref{tab:verbosity_pilot_study}, all models tested exhibit low VT rates, indicating that a substantial portion of their computational effort (often 35-45\%) is spent on redundant reasoning after the solution has been found. This widespread inefficiency confirms the significant potential for compression and motivates our work.

\subsection{Principles for Efficient Reasoning}
\label{principle}
The evaluation of reasoning models traditionally rests on two pillars: \textit{Efficiency} (the computational cost, often proxied by output length) and \textit{Efficacy} (the ability to solve the problem correctly). However, simply shortening the output is a coarse approach that may inadvertently remove critical thinking steps. To create a more targeted framework, we refine these concepts by introducing two new, complementary principles:

\begin{itemize}[leftmargin=*, itemsep=0pt]

\item \textbf{\textit{Brevity}} refines Efficiency by shifting the focus from generic length reduction to the specific elimination of ``\textit{Redundant Sequence}''. While conventional methods may still produce a compressed sequence that contains unnecessary double-checks, Brevity advocates for the model to terminate its reasoning process as soon as the correct answer is found.

\item \textbf{\textit{Sufficiency}} acts as a crucial safeguard for Efficacy. It mandates that, in the pursuit of Brevity, no critical logical steps essential for reaching a correct answer are omitted. It ensures that the compressed reasoning remains complete and logically sound.

\end{itemize}

Therefore, the ideal reasoning model must navigate the tension between these principles: it should be maximally \textit{Brief} by removing all non-essential thinking, yet always remain \textit{Sufficient} to guarantee correctness. Our work, \method, is explicitly designed to optimize for this balance.

\begin{figure*}[!t]
    \centering
    \includegraphics[width=0.9\textwidth]{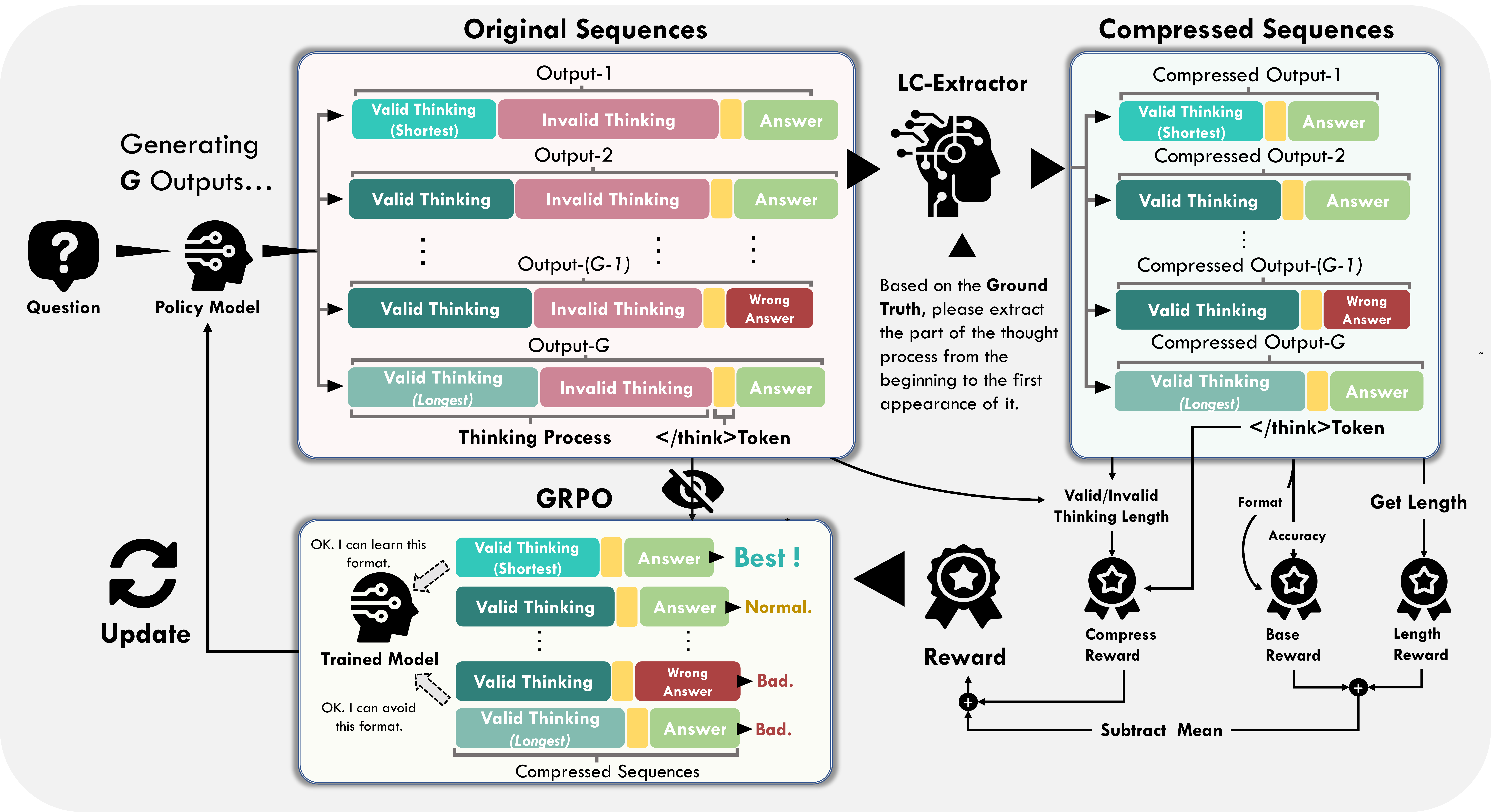}
    \caption{\textbf{An overview of the \method training three-stage pipeline.} \textbf{\textit{(1)} Valid Segment Extraction:} First, an extractor model processes the original reasoning traces to identify the valid thinking portion and generate compressed sequences. \textbf{\textit{(2)} Reward Calculation:} Next, these compressed sequences are used to compute our dual rewards—Length Reward and Compress Reward, with the latter applied exclusively as a bonus or penalty on the final \texttt{</think>} token. These are then combined to calculate the final advantages. \textbf{\textit{(3)} Policy Optimization:} Finally, the GRPO loss is calculated using the compressed sequences and corresponding advantages, steering the model toward more concise and efficient reasoning. }
    \label{fig:grpo}

\end{figure*}

\section{\method: Length Compression with Efficient Reasoning Principles}
In this section, we propose \method, a GRPO-based post-training algorithm designed to address the ``\textit{invalid thinking}'' phenomenon and enhance reasoning efficiency. Guided by the principles of \textit{Brevity} and \textit{Sufficiency} introduced in Section \ref{principle}, \method employs a novel dual-reward system. This system combines a global Length Reward for overall conciseness with a targeted Compress Reward that specifically removes redundant reasoning. The complete pipeline of \method is illustrated in Figure \ref{fig:grpo} and Algorithm \ref{alg:lcr1_concise}.
\subsection{Problem Formulation}

Let $\mathcal M$ be the model and $q$ be the given query. The output is $o\sim\mathcal M(q)$, where $o=\text{cat}(R,A)$ consists of a reasoning part $R$ and an answer part $A$, split by the token \texttt{</think>}, which is considered part of $A$. For the reasoning part $R$, we denote its effective prefix $R'$  as the content from the beginning of $R$ up to the first occurrence of the correct answer corresponding to the query $q$. If $R$ does not contain the correct answer, then we define $R'=R$. We define two functions as follows:
\begin{equation}
\label{Eq: notation}
    t(\{R,A\})=R,\quad f(\{R,A\})=\{R',A\}
\end{equation}
The function $t$ extracts the reasoning process $R$ from the output $o$ and function $f$ extracts the concise reasoning part $R'$ and concatenates it with the answer $A$. We denote $o_i$ as the original model output and $o_i'=f(o_i)$ as the refined compressed output.

\method is a GRPO-based method to efficiently compress the reasoning process. Within a group, let $\mathcal{C}$ denote the set of indices $i$ where sequences $o_i$ leading to the correct answer corresponding to the query $q$, and $\mathcal{W}$ be the set of indices $j$ where $o_j$ leading to a wrong answer, The total group size is $G = |\mathcal{C}| + |\mathcal{W}|$.

\begin{algorithm*}[htbp]
\caption{\method: \textbf{L}ength \textbf{C}ompress for \textbf{R1}-style model}
\label{alg:lcr1_concise}
\textbf{Input:} Initial policy model $\pi_{\theta}$, compression function $f(\cdot)$, task prompts $\mathcal{D}$, hyperparameters $\alpha, \beta, \mu$ \\
\textbf{Output:} Trained policy model $\pi_\theta$
\begin{algorithmic}[1]
    \FOR{step = $1, \dots, M$}
        \STATE Sample a batch $\mathcal{D}_b$ from $\mathcal{D}$
        \STATE Update the old policy model $\pi_{\theta_{\text{old}}} \leftarrow \pi_\theta$
        \STATE Sample $G$ outputs $\{o_i\}_{i=1}^G \sim \pi_{\theta_{\text{old}}}(\cdot|q)$ for each question $q \in \mathcal{D}_b$
        \STATE Apply compression to all outputs: $o_i' \leftarrow f(o_i)$
        \STATE Compute combined reward $r_{i,\text{combine}}$ (Eq. \ref{eq:base_reward}) and compress reward $r_{i,\text{compress}}$ (Eq. \ref{eq:compression_reward})
        \STATE Compute token-level advantages $\hat{A}_{i,t}$ for each compressed output $o_i'$ (Eq. \ref{eq:advantage})
        \FOR{iteration = $1, \dots, \mu$}
            \STATE Update the policy model $\pi_\theta$ by maximizing the objective $\mathcal{J}_{\text{GRPO}}$ (Eq. \ref{eq:GRPO})
        \ENDFOR
    \ENDFOR
    \STATE \textbf{return} $\pi_\theta$
\end{algorithmic}
\end{algorithm*}

\subsection{Reward and Objective Design}
\label{sec:reward_equations}
Our method's reward system consists of two core components: the Length Reward for reducing overall output length, and the Compress Reward for targeting redundant parts of the model's reasoning.
\paragraph{Length Reward.} To compress the total length of the model output, we propose adding a length penalty during the GRPO training process. Leveraging the group-based sampling of GRPO, we can calculate the relative length reward to automatically adjust to the difficulty of the problem. And we define the Length Reward as follows:
\begin{equation}
    r_{i,\text{length}} =
    \begin{cases}
        1 - \frac{|o_i'|}{\max_{j \in \mathcal{C}} |o_j'|}, & \text{if } i \in \mathcal{C} \\
        0, & \text{if } i \in \mathcal{W}
    \end{cases}
    \label{eq:length_reward}
\end{equation}
This formulation uses the maximum length of a correct, compressed sequence within the group as a normalizer. The final reward combines this with a base reward for format and accuracy, and is normalized by subtracting the group mean, following \citet{liu2025drgrpo} to obtain an unbiased gradient:
\begin{gather}
    \tilde{r}_i = r_{i,\text{base}} + \alpha \cdot r_{i,\text{length}} \\
    r_{i,\text{combine}} = \tilde{r}_i - \operatorname{mean}(\{\tilde{r}_j\}_{j=1}^G)
    \label{eq:base_reward}
\end{gather}
where
\begin{equation}
    r_{i,\text{base}} = r_{i,\text{format}} + r_{i,\text{accuracy}} \\ 
\end{equation}
Following prior work, $r_{i,\text{format}}$ and $r_{i,\text{accuracy}}$ are binary rewards to judge whether the model places its thinking process between \texttt{<think>} and \texttt{</think>} and whether the sample leads to the correct answer corresponding to the query verified by Math-Verify\footnote{https://github.com/huggingface/Math-Verify} respectively. $\alpha$ is a hyperparameter that controls the weight of the Length Reward.
\paragraph{Compress Reward.} For the original GRPO method, the loss calculation is based on the model's own sampling results. In order to drive the model to terminate the thinking process when getting the correct answer for achieving Brevity, we modify the GRPO objective as follows:
\begin{align}
&\mathcal{J}_{\text{GRPO}}(\theta) = \mathbb{E}_{q \sim P(Q), \{o_i\}_{i=1}^G \sim \pi_{\theta_{\text{old}}}(O|q)}  \label{eq:GRPO}\\
    & \!\!\Biggl[\ \frac{1}{\sum_{i=1}^G |o'_i|} 
       \sum_{i=1}^G \sum_{t=1}^{|o'_i|}
       \Biggl\{\! 
         \min\!\bigl[R_t(\theta)\cdot \hat{A}_{i,t},\; \operatorname{clip}\!\Bigl(
         R_t(\theta),\nonumber \\
    & 
         \,1-\epsilon,\;1+\epsilon
       \Bigr)\cdot \hat{A}_{i,t}\bigr] -\beta \, 
         D_{\text{KL}}\!\bigl(
           \pi_\theta(\cdot|q)\,\|\,\pi_{\text{ref}}(\cdot|q)
         \bigr)
       \Biggr\} \Biggr] \nonumber
\end{align}
where 
\begin{equation}
    \mathbb{D}_{\text{KL}}\!\bigl(
           \pi_\theta\,\|\,\pi_{ref}
         \bigr)=\frac{\pi_{ref}(o_i'|q)}{\pi_\theta(o_i'|q)}-\log \frac{\pi_{ref}(o_i'|q)}{\pi_\theta(o_i'|q)} -1
\end{equation}
\begin{equation}
    o_i' = f(o_i),\quad R_t(\theta) = \frac{\pi_\theta(o'_{i,t}|q, o'_{i,<t})}{\pi_{\theta_{\text{old}}}(o'_{i,t}|q, o'_{i,<t})}
    \label{eq:prob_ratio}
\end{equation}
Our key modification to the standard GRPO objective is that the loss is calculated over the compressed trajectories $o'_i$, rather than the original full trajectories $o_i$. We define the token-level advantages $\hat A_{i,t}$ as follows:
\begin{equation}
    \hat{A}_{i,t} = r_{i,\text{combine}} +\gamma \cdot \mathbb I(o_{i,t}'=\texttt{</think>})\cdot r_{i,\text{compress}}
    \label{eq:advantage}
\end{equation}
where
\begin{equation}
    r_{i,\text{compress}} = 
    \begin{cases}
        1 - \frac{|t(o_i')|}{|t(o_i)|}, & \text{if } i \in \mathcal{C} \text{ \& ans}(q) \in t(o'_i) \\
        -1, & \text{if } i \in \mathcal{C} \text{ \& ans}(q) \notin t(o'_i) \\
        0, & \text{if } i \in \mathcal{W}
    \end{cases}
    \label{eq:compression_reward}
\end{equation}
Let $\text{ans}(q)$ be the ground truth answer for a given query $q$. In this setting, we place focuses on steering model towards outputting \texttt{</think>} token when getting the correct answer (at the end of $o'_i$) during the thinking process to achieve compressing the verbose token, conforming to the principle of \textbf{Brevity}. We only give an extra reward to this token, avoiding place unnecessary emphasis on other tokens to make the training process more efficient and stable. We define the reward to be the portion of Redundant Sequence, formulated by $1 - \frac{|t(o_i')|}{|t(o_i)|}$, representing the efficiency distinction between sequences before and after compression. The hyperparameter 
$\gamma$ scales this bonus.

Based on the principle of \textbf{Sufficiency}, the model should engage in sufficient reasoning process, avoiding overhead compression at the cost of accuracy degradation. Therefore, we impose a large penalty (-1) to the token \texttt{</think>} if the model terminates its reasoning before finding the correct answer,  which discourages harmful over-compression and provides robustness to the training process.

To further validate the effectiveness of our method, we follow DAPO \citep{yu2025dapo} to  calculate the objection across all tokens in a group, instead of averaging the token rewards within a single sequence, which eliminates the original GRPO method's preference for short-correct sequences and long-incorrect sequences, facilitating the validation of our method's effectiveness.

\section{Experiments}\label{experiment}

\begin{table*}[!t]
 \centering
 \small
 \setlength{\tabcolsep}{3pt}
 \renewcommand{\arraystretch}{1.0}
 \newcolumntype{C}[1]{>{\centering\arraybackslash}p{#1}}
 \caption{Accuracy (above) and Sequence Length (below) for all methods across seven benchmarks. \textit{AVG} shows the relative change in accuracy and length compared to the Base model (+ increase, - decrease). GPQA-D denotes GPQA-Diamond benchmark, and LCB denotes the pass@10 score on LiveCodeBench. \textit{VT} represents the Valid Thinking ratio. For each column, the best performing score is marked in \textbf{bold}, and the \underline{second-best} is underlined.}

 \label{tab:2}
 \resizebox{0.9\textwidth}{!}{
 \begin{tabular}{l *{8}{C{1.25cm}} C{1.5cm}}
 \toprule
 \textbf{Method} & \textbf{AIME25} & \textbf{MATH500} & \textbf{GSM8K} & \textbf{Olympiad} & \textbf{AMC} &\textbf{GPQA-D} &\textbf{LCB} & \textbf{Avg} & \textbf{VT}\\
 \midrule
 \multicolumn{10}{c}{\textbf{DeepSeek-R1-Distill-Qwen-7B}} \\
 \midrule
 \multirow{2}{*}{Base}
  & 37.7 & 92.6 & 91.6 & 59.7 & 81.2 &46.6&68.8& \multirow{2}{*}{–}& \multirow{2}{*}{58.72\%} \\
  & (11007) & (3832) & (1842) & (7342) & (6715) &(6508)&(8878)& &\\
 \cmidrule(lr){1-10}
 \multirow{2}{*}{SFT}
  & 36.6 & 90.2 & \textbf{91.9} & 56.0 & 78.7 &39.8&67.3& –4.46\% &\multirow{2}{*}{95.64\%}\\
  & (9457) & (2497) & (946) & (6329) & (5231) &(8217)&(8739)& (–15.54\%) &\\
 \cmidrule(lr){1-10}
 \multirow{2}{*}{DPO}
  & 36.9 & 91.4 & 90.3 & 56.2 & 78.6 &37.2&66.9& –5.26\% & \multirow{2}{*}{\underline{96.34\%}}\\
  & (9718) & (2277) & (980) & (6338) & (5122) &(8109)&(8755)& (–16.18\%) & \\
 \cmidrule(lr){1-10}
 \multirow{2}{*}{O1-Pruner}
  & 35.0 & 91.5 & 91.1 & 59.6 & 77.1 &45.5&66.7& –2.79\% & \multirow{2}{*}{69.30\%}\\
  & (8263) & (2268) & (1012) & (4712) & (4510) &(5012)&(\textbf{5901})& (\underline{–33.71\%})& \\
 \cmidrule(lr){1-10}
 \multirow{2}{*}{ThinkPrune}
  & \textbf{38.0} & \textbf{93.1} & 91.2 & \textbf{60.8} & \textbf{82.7} &\textbf{50.3}&67.8& \textbf{+1.58\%} & \multirow{2}{*}{77.16\%}\\
  & (9309) & (3253) & (1546) & (6225) & (5510) &(6508)&(7180)& (–14.13\%)& \\
 \cmidrule(lr){1-10}
 \multirow{2}{*}{SFT+O1-Pruner}
  & 35.5 & 91.0 & 89.7 & 56.0 & 76.6 &43.9&66.8& –4.31\% & \multirow{2}{*}{85.22\%}\\
  & (9466) & (2245) & (920) & (5807) & (5133) &(6425)&(7267)& (–24.19\%)& \\
 \cmidrule(lr){1-10}
 \multirow{2}{*}{\textbf{\method (Ours)}}
  & 36.2 & 90.4 & 88.1 & 58.7 & 79.1 &47.2&\textbf{69.0}& \underline{–1.84\%} & \multirow{2}{*}{\textbf{97.14\%}}\\
  & (\textbf{7150}) & (\textbf{1568}) & (\textbf{450}) & (\textbf{4041}) & (\textbf{3453}) &(\textbf{4604})&(6059)& (\textbf{–46.32\%}) & \\
 \midrule

 \multicolumn{10}{c}{\textbf{DeepSeek-R1-Distill-Qwen-1.5B}}  \\
 \midrule
 \multirow{2}{*}{Base}
  & 22.8 & 83.7 & 83.4 & 44.2 & 61.2 &\textbf{34.5}&\textbf{43.1}& \multirow{2}{*}{–} &\multirow{2}{*}{56.06\%} \\
  & (12129) & (4869) & (2294) & (9258) & (8696) &(8516)&(10120)& & \\
 \cmidrule(lr){1-10}
 \multirow{2}{*}{SFT}
  & 20.5 & 81.4 & 81.3 & 42.7 & 59.7 &22.4&39.8& –9.13\% &\multirow{2}{*}{95.54\%} \\
  & (10639) & (3045) & (1134) & (7637) & (6608) &(10217)&(10597)& (–16.74\%) & \\
 \cmidrule(lr){1-10}
 \multirow{2}{*}{DPO}
  & 19.4 & 79.0 & 80.9 & 41.1 & 56.7 &19.8&39.2& –12.79\% & \multirow{2}{*}{\underline{98.38\%}}\\
  & (10316) & (2749) & (855) & (6544) & (5912) &(9438)&(10287)& (–24.30\%) & \\
 \cmidrule(lr){1-10}
 \multirow{2}{*}{O1-Pruner}
  & 23.2 & 84.3 & 82.7 & \textbf{47.1} & \textbf{65.1} &32.1&42.5& \underline{+0.89\%} & \multirow{2}{*}{78.20\%}\\
  & (8731) & (2913) & (1162) & (5960) & (5131) &(6173)&(7305)& (\underline{–35.64\%}) & \\
 \cmidrule(lr){1-10}
 \multirow{2}{*}{ThinkPrune}
  & \textbf{24.1} & \textbf{84.5} & \textbf{84.1} & 44.9 & 63.4 &33.6&42.7& \textbf{+1.31\%} & \multirow{2}{*}{65.62\%}\\
  & (7960) & (3518) & (1690) & (6250) & (5897) &(5576)&(7226)& (–30.89\%) & \\
 \cmidrule(lr){1-10}
 \multirow{2}{*}{SFT+O1-Pruner}
  & 17.5 & 80.2 & 81.5 & 40.0 & 58.7 &25.0&39.4& –11.34\% & \multirow{2}{*}{91.38\%}\\
  & (9075) & (2769) & (919) & (6411) & (5553) &(7410)&(8488)& (–32.04\%) & \\
 \cmidrule(lr){1-10}
 \multirow{2}{*}{\textbf{\method (Ours)}}
  & 21.2 & 82.5 & 82.7 & 43.2 & 61.7 &33.6&42.4& –2.14\% & \multirow{2}{*}{\textbf{98.64\%}}\\
  & (\textbf{6434}) & (\textbf{2233}) & (\textbf{841}) & (\textbf{4333}) & (\textbf{3947}) &(\textbf{4489})&(\textbf{5722})& (\textbf{–51.86\%}) & \\
 \bottomrule
 \end{tabular}
 }
 \end{table*}

\begin{table*}[!t]
\centering
\small
\setlength{\tabcolsep}{3pt}
\renewcommand{\arraystretch}{1.0}
\newcolumntype{C}[1]{>{\centering\arraybackslash}p{#1}}
\caption{Ablation study on the contribution of Length Reward and Compress Reward to the compression process. The study manifest the sub-optimal performance of them, varifying each of them makes a big contribution to the efficient reasoning. } 
\label{tab:exp2}
\resizebox{0.9\textwidth}{!}{
\begin{tabular}{l *{8}{C{1.25cm}} C{1.5cm}}
\toprule
\textbf{Method} & \textbf{AIME25} & \textbf{MATH500} & \textbf{GSM8K} & \textbf{Olympiad} & \textbf{AMC} & \textbf{GPQA-D} & \textbf{LCB} & \textbf{Avg} & \textbf{VT} \\
\midrule
\multicolumn{10}{c}{\textbf{DeepSeek-R1-Distill-Qwen-7B}} \\
\midrule
\multirow{2}{*}{\textbf{\method (Ours)}}
  & 36.2 & 90.4 & 88.1 & 58.7 & 79.1 &47.2&\textbf{69.0}& –1.84\% & \multirow{2}{*}{\textbf{97.14\%}}\\
  & (\textbf{7150}) & (\textbf{1568}) & (\textbf{450}) & (\textbf{4041}) & (\textbf{3453}) &(\textbf{4604})&(\textbf{6059})& (\textbf{–46.32\%}) & \\
\cmidrule(lr){1-10}
\multirow{2}{*}{w/o L-reward}
  & 36.1 & 91.3 & 90.6 & \textbf{59.4} & 79.0 & 45.9&68.0 & \underline{–1.80\%} & \multirow{2}{*}{\underline{93.16\%}}\\
  & (9309) & (2316) & (696) & (5779) & (5021) &(6273) &(8023) & (–25.28\%) & \\
\cmidrule(lr){1-10}
\multirow{2}{*}{w/o C-reward}
  & \textbf{37.6} & \textbf{92.9} & \textbf{91.1} & 59.1 & \textbf{80.5} &\textbf{48.9} &68.5 & \textbf{+0.31\%} & \multirow{2}{*}{72.24\%}\\
  & (8738) & (2498) & (1012) & (5344) & (4741) & (5727)&(6893) & (\underline{–27.35\%}) & \\
\midrule
\multicolumn{10}{c}{\textbf{DeepSeek-R1-Distill-Qwen-1.5B}} \\
\midrule
\multirow{2}{*}{\textbf{\method (Ours)}}
  & 21.2 & 82.5 & 82.7 & 43.2 & 61.7 &\textbf{33.6}&42.4& \underline{–2.14\%} & \multirow{2}{*}{\textbf{98.64\%}}\\
  & (\textbf{6434}) & (\textbf{2233}) & (841) & (\textbf{4333}) & (\textbf{3947}) &(\textbf{4489})&(\textbf{5722})& (\textbf{–51.86\%}) & \\
\cmidrule(lr){1-10}
\multirow{2}{*}{w/o L-reward}
  & 21.3 & 81.2 & 83.3 & 43.4 & 62.2 &  30.6&41.9 & –3.42\% &\multirow{2}{*}{\underline{95.16\%}} \\
  & (7061) & (2270) & (\textbf{754}) & (5024) & (4478) & (5021)&(6378) & (\underline{–47.79\%}) & \\
\cmidrule(lr){1-10}
\multirow{2}{*}{w/o C-reward}
  & \textbf{21.9} & \textbf{83.2} & \textbf{84.1} & \textbf{44.0} & \textbf{63.4} & 30.1 &\textbf{43.7} & \textbf{–1.70\%} & \multirow{2}{*}{71.10\%}\\
  & (7988) & (2965) & (1160) & (5363) & (5192) &(5847) &(6874) & (–38.35\%) & \\
\bottomrule
\end{tabular}
}
\end{table*}
\subsection{Experiment Setups}

\paragraph{Backbone Models.}
We choose two representative reasoning models: DeepSeek-R1-Distill-Qwen-7B/1.5B \citep{deepseekai2025deepseek-r1} as our backbone models, which have demonstrated strong performance on mathematical and coding reasoning tasks.

\paragraph{\extractor.}
To accurately identify and extract the valid reasoning part, we develop a specialized parser to implement the extraction function $f$ mentioned in Eq. \ref{Eq: notation}, termed \extractor. We finetune Qwen2.5-3B-Instruct for its lightweight and easy enough to run. Detailed experiment settings are provided in Appendix \ref{Appendix: LC-Extractor}.

\paragraph{Dataset.}
We used a mixed-difficulty dataset, combining past AIME competition problems with the MATH dataset in an approximate 1:2 ratio to create 2500 training samples. This approach enables the model to learn length compression across problems of varying difficulty.

\paragraph{Evaluation.}
We test our model's performance on seven datasets, including AIME25, MATH500, GSM8K, AMC, OlympiadBench, GPQA-Diamond and LiveCodeBench, across math, general and code tasks, to evaluate the efficiency of reasoning comprehensively. We use averaged Pass@1 as our primary metric. For each test, we sample N times, setting top-p $=0.95$ and temperature $=0.7$. For AIME25, we set $N=64$, while for the other test sets, we set $N=8$. We set the maximum length to $16384$. Additionally, we calculate the average fluctuate ratio on accuracy and token lengths compared with base model on every benchmark, which can be formulated as follows:
\begin{align}
\label{metric}
\text{Avg}_{\text{acc}} &= \text{mean}_{i=1}^7\Big\{\frac{\text{Acc}^{\text{model}}_i-\text{Acc}^{\text{base}}_i}{\text{Acc}^{\text{base}}_i}\Big\} \\
\text{Avg}_{\text{len}} &= \text{mean}_{i=1}^7\Big\{\frac{\text{Len}^{\text{model}}_i-\text{Len}^{\text{base}}_i}{\text{Len}^{\text{base}}_i}\Big\}
\end{align}
We also test VT for each model to evaluate the Brevity of the thinking process to investigate the ability of these methods to mitigate the ``invalid thinking'' phenomenon. We test VT on five math benchmarks and calculate the mean value, for the convenience of extracting the standard and formatted correct answer from the thinking process on math problems.

\subsection{Baselines}
\paragraph{Supervised Fine-tuning (SFT).} Inspired by \textsc{OverThink} \citep{chen2024overthinking}, which proposes using only the initial correct solution for fine-tuning, we construct an SFT dataset of 5000 samples by removing the Redundant Sequence from self-generated outputs.

\paragraph{Direct Preference Optimization (DPO) \citep{rafailov2023direct}.} We create a preference dataset of 5000 samples from the MATH dataset, where the shortest correct answer is treated as the \textit{``chosen''} response and the longest as the \textit{``rejected''} response. This DPO training is applied to the SFT-tuned model.

\paragraph{O1 Pruner \citep{luo2025o1prunerlengthharmonizingfinetuningo1like}.}  A PPO-like offline fine-tuning method to significantly compress CoT length while maintaining performance. We follow its methodology using 10000 samples from the MATH dataset.

\paragraph{ThinkPrune-3K \citep{hou2025thinkprune}.} A reinforcement learning approach that uses a length-truncation reward for multi-stage compression. We reproduce the ThinkPrune-3k variant, which is reported to be highly efficient, with slight accuracy degradation.

\paragraph{SFT + O1-Pruner.} To better understand the effect of compressing the thinking process and pruning the overall sequences at the same time, we also compare with a two-stage training approach, combining SFT and O1 Pruner.

\subsection{Experiment Results}
 
\paragraph{\method outperforms other methods with competitive performance and fewer tokens.} As presented in Table \ref{tab:2}, On the 7B model, \method achieves an average length reduction of 46.32\%, substantially higher than all other baselines, with a mere 1.84\% drop in average accuracy. Similarly, on the 1.5B model, it attains a 51.86\% length reduction for a 2.14\% accuracy decrease. 
This efficiency does not appear to compromise its generalization, as it demonstrates more robust performance on out-of-distribution (OOD) benchmarks like GPQA-Diamond and LiveCodeBench compared to other high-compression methods. Figure \ref{fig:pareto} shows our method achieves more favorable Efficacy-Efficiency trade-off by enabling maximal compression ratio with negligible accuracy degradation.
\method also achieves a significantly higher VT rate (over 97\%) compared to other methods like O1-Pruner (\textasciitilde 70-78\%) and ThinkPrune (\textasciitilde 66-77\%), demonstrating the superior efficiency of our approach.

\begin{figure*}[!t]
    \centering
    \includegraphics[width=1\linewidth]{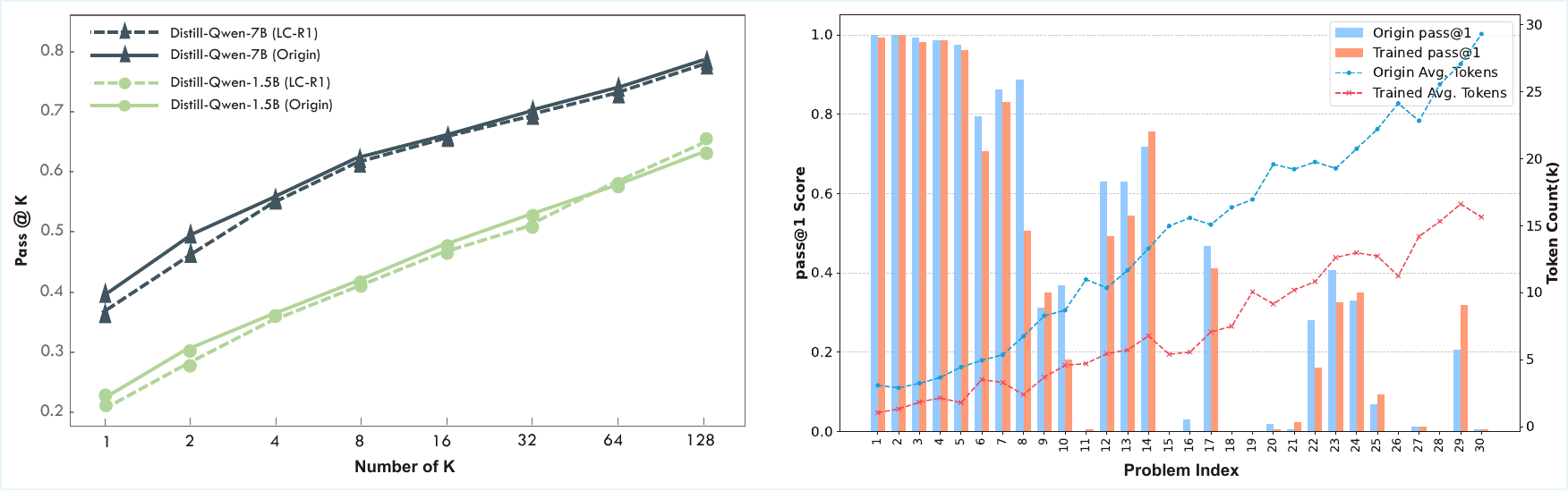}
    \vspace{-1em}
    \caption{The impact of \method compression method on the AIME25 benchmark. \textbf{\textit{Left:}} The Pass@k scores show that \method models maintain competitive performance compared to the originals, preserving the model's potential. \textbf{\textit{Right:}} Per-problem analysis on Deepseek-R1-Distill-Qwen-7B reveals that \method achieves similar Pass@1 accuracy while maintaining a consistent token compression ratio across problems of varying difficulty, demonstrating a universal compression effect.}
    \label{fig:combine}
    \vspace{-1em}
\end{figure*}

\paragraph{Combining length and compress reward brings superior efficiency to reasoning.} 
Our ablation study on the Length Reward (L-reward) and Compress Reward (C-reward), presented in Table \ref{tab:exp2}, reveals their critical complementary relationship. The analysis reveals that while each component alone yields competitive results—positioning them near the Pareto frontier of performance versus compression efficiency—combining them can achieve a more optimal balance. Specifically, using L-reward alone achieves significant compression but with lower VT rate. Conversely, C-reward alone ensures a high VT by precisely removing redundancy, but with limited overall compression. Our full \method method successfully integrates these strengths, achieving both the highest compression efficiency and the highest VT rate while maintaining comparable accuracy, proving that the synergy between both rewards is indispensable for achieving maximum reasoning efficiency.

\paragraph{SFT shows limitation on generalization.} While SFT achieves a remarkably high VT rate (over 95\%), its effectiveness is superficial. The model's performance collapses on \textit{OOD} benchmarks, indicating that it merely overfits to the structural brevity of the training data rather than learning a generalizable, efficient reasoning policy. The poor performance of the hybrid SFT+O1-Pruner method further suggests that a simple combination of off-the-shelf techniques is insufficient. These findings underscore the superiority of RL-based methods like \method, which foster more robust and genuinely efficient reasoning skills.

\section{Compression Impact Analysis}

\paragraph{Compression does not impact exploration capability.} To investigate the deeper impact of compression on the model's problem-solving potential, we sampled 256 times on AIME25 with maximal length $=32,768$ and test pass@k score on both models before and after compression. The results in Figure \ref{fig:combine} (left) reveal a key phenomenon: across the entire Pass@k evaluation range from k $=1$ to $128$ on the AIME25 dataset, the performance curve of the model compressed by our \method method almost perfectly overlaps with that of the original model. This result strongly demonstrates that  the model's exploration ability to find a correct solution through multiple attempts will not be injured by the compression. It suggests that the pruned ``invalid thinking'' segments are truly redundant and their removal does not diminish the model's underlying knowledge or creative problem-solving potential.

\paragraph{Compression remains consistent across varying problem difficulties.} To analyze our method's behavior at a microscopic level, we plot the per-problem pass@1 accuracy against the original model's token consumption on the AIME25 benchmark (Figure \ref{fig:combine} (right)). The plot reveals a clear difficulty spectrum, where problems requiring more tokens from the base model generally correspond to lower pass@1 scores. Crucially, \method applies a uniform and significant compression ratio across this entire spectrum, with per-problem outcomes (\emph{i.e.}, success or failure) remaining remarkably consistent with those of the base model. This provides strong evidence that \method functions as a robust and difficulty-agnostic efficiency layer, successfully streamlining the reasoning process without altering the model's core problem-solving logic for any specific problem.

\section{Conclusion}
In this paper, we address the ``\textit{invalid thinking}'' phenomenon existing in current LRMs, that they tend to double-check their work after correct answer has been derived. To tackle this, we introduced the principles of Brevity and Sufficiency and proposed \method, a RL-based post-training method that employs a dual-reward system, compressing overall sequence length and pruning Redundant Sequence spontaneously.
Extensive experiments demonstrate that \method achieves more favorable \textit{Efficacy-Efficiency} trade-off. In further analysis, \method does not degrade the model's exploration ability and and the compression effect remains robust  across problems of varying difficulty.

\section*{Impact Statement}
In this paper, we address the ``invalid thinking'' phenomenon, a key source of inefficiency in Large Reasoning Models where they engage in unnecessary verification after deriving a correct answer. We introduce \method, a novel post-training method featuring a dual-reward system that encourages both overall conciseness and the specific elimination of this redundancy. Our experiments demonstrate that \method achieves a more favorable trade-off between performance and efficiency than existing approaches. While our current validation focuses on models up to the 7B scale due to computational constraints, this work provides a proven path toward developing more computationally frugal LRMs. By making advanced AI reasoning more efficient, we hope to make these powerful tools more scalable and accessible for a wider range of applications.

\section*{Acknowledgment }
Many thanks to Yao Wan and for his invaluable support and comments.

{
    \small
    \bibliographystyle{unsrtnat}
    \bibliography{custom}

\begin{thebibliography}{42}
\providecommand{\natexlab}[1]{#1}
\providecommand{\url}[1]{\texttt{#1}}
\expandafter\ifx\csname urlstyle\endcsname\relax
  \providecommand{\doi}[1]{doi: #1}\else
  \providecommand{\doi}{doi: \begingroup \urlstyle{rm}\Url}\fi

\bibitem[Jaech et~al.(2024)Jaech, Kalai, Lerer, Richardson, El-Kishky, Low, Helyar, Madry, Beutel, Carney, et~al.]{jaech2024openai}
Aaron Jaech, Adam Kalai, Adam Lerer, Adam Richardson, Ahmed El-Kishky, Aiden Low, Alec Helyar, Aleksander Madry, Alex Beutel, Alex Carney, et~al.
\newblock Openai o1 system card.
\newblock \emph{arXiv preprint arXiv:2412.16720}, 2024.

\bibitem[DeepSeek-AI et~al.(2025)DeepSeek-AI, Guo, Yang, Zhang, Song, Zhang, Xu, Zhu, Ma, Wang, Bi, Zhang, Yu, Wu, Wu, Gou, Shao, Li, ..., and Zhang]{deepseekai2025deepseek-r1}
DeepSeek-AI, Daya Guo, Dejian Yang, Haowei Zhang, Junxiao Song, Ruoyu Zhang, Runxin Xu, Qihao Zhu, Shirong Ma, Peiyi Wang, Xiao Bi, Xiaokang Zhang, Xingkai Yu, Yu~Wu, Z.~F. Wu, Zhibin Gou, Zhihong Shao, Zhuoshu Li, Ziyi~Gao ..., and Zhen Zhang.
\newblock Deepseek-r1: Incentivizing reasoning capability in llms via reinforcement learning, 2025.
\newblock URL \url{https://arxiv.org/abs/2501.12948}.

\bibitem[Wei et~al.(2023)Wei, Wang, Schuurmans, Bosma, Ichter, Xia, Chi, Le, and Zhou]{wei2023chain-of-thought}
Jason Wei, Xuezhi Wang, Dale Schuurmans, Maarten Bosma, Brian Ichter, Fei Xia, Ed~Chi, Quoc Le, and Denny Zhou.
\newblock Chain-of-thought prompting elicits reasoning in large language models, 2023.
\newblock URL \url{https://arxiv.org/abs/2201.11903}.

\bibitem[Sun et~al.(2025)Sun, Min, Chen, Zhao, Liu, Wang, Fang, and Wen]{sun2025olym-math}
Haoxiang Sun, Yingqian Min, Zhipeng Chen, Wayne~Xin Zhao, Zheng Liu, Zhongyuan Wang, Lei Fang, and Ji-Rong Wen.
\newblock Challenging the boundaries of reasoning: An olympiad-level math benchmark for large language models, 2025.
\newblock URL \url{https://arxiv.org/abs/2503.21380}.

\bibitem[Gu et~al.(2024)Gu, Rozière, Leather, Solar-Lezama, Synnaeve, and Wang]{gu2024cruxeval}
Alex Gu, Baptiste Rozière, Hugh Leather, Armando Solar-Lezama, Gabriel Synnaeve, and Sida~I. Wang.
\newblock Cruxeval: A benchmark for code reasoning, understanding and execution.
\newblock \emph{arXiv preprint arXiv:2401.03065}, 2024.

\bibitem[Chen et~al.(2025)Chen, Qin, Liu, Peng, Guan, Wang, Hu, Zhou, Gao, and Che]{chen2025reasoning-era-survey-long}
Qiguang Chen, Libo Qin, Jinhao Liu, Dengyun Peng, Jiannan Guan, Peng Wang, Mengkang Hu, Yuhang Zhou, Te~Gao, and Wanxiang Che.
\newblock Towards reasoning era: A survey of long chain-of-thought for reasoning large language models, 2025.
\newblock URL \url{https://arxiv.org/abs/2503.09567}.

\bibitem[Aggarwal and Welleck(2025)]{aggarwal2025l1controllinglongreasoning}
Pranjal Aggarwal and Sean Welleck.
\newblock L1: Controlling how long a reasoning model thinks with reinforcement learning, 2025.
\newblock URL \url{https://arxiv.org/abs/2503.04697}.

\bibitem[Chen et~al.(2024)Chen, Xu, Liang, He, Pang, Yu, Song, Liu, Zhou, Zhang, et~al.]{chen2024overthinking}
Xingyu Chen, Jiahao Xu, Tian Liang, Zhiwei He, Jianhui Pang, Dian Yu, Linfeng Song, Qiuzhi Liu, Mengfei Zhou, Zhuosheng Zhang, et~al.
\newblock Do not think that much for 2+ 3=? on the overthinking of o1-like llms.
\newblock \emph{arXiv preprint arXiv:2412.21187}, 2024.

\bibitem[Sui et~al.(2025)Sui, Chuang, Wang, Zhang, Zhang, Yuan, Liu, Wen, Chen, Hu, et~al.]{sui2025stop-overthinking-survey}
Yang Sui, Yu-Neng Chuang, Guanchu Wang, Jiamu Zhang, Tianyi Zhang, Jiayi Yuan, Hongyi Liu, Andrew Wen, Hanjie Chen, Xia Hu, et~al.
\newblock Stop overthinking: A survey on efficient reasoning for large language models.
\newblock \emph{arXiv preprint arXiv:2503.16419}, 2025.

\bibitem[Cuadron et~al.(2025)Cuadron, Li, Ma, Wang, Wang, Zhuang, Liu, Schroeder, Xia, Mao, et~al.]{cuadron2025overthinking-danger}
Alejandro Cuadron, Dacheng Li, Wenjie Ma, Xingyao Wang, Yichuan Wang, Siyuan Zhuang, Shu Liu, Luis~Gaspar Schroeder, Tian Xia, Huanzhi Mao, et~al.
\newblock The danger of overthinking: Examining the reasoning-action dilemma in agentic tasks.
\newblock \emph{arXiv preprint arXiv:2502.08235}, 2025.

\bibitem[Luo et~al.(2025{\natexlab{a}})Luo, He, Wang, Yang, Liu, Tan, Cao, Tao, and Shen]{luo2025adar1}
Haotian Luo, Haiying He, Yibo Wang, Jinluan Yang, Rui Liu, Naiqiang Tan, Xiaochun Cao, Dacheng Tao, and Li~Shen.
\newblock Adar1: From long-cot to hybrid-cot via bi-level adaptive reasoning optimization.
\newblock \emph{arXiv preprint arXiv:2504.21659}, 2025{\natexlab{a}}.

\bibitem[Shen et~al.(2025)Shen, Zhang, Huang, Shi, Zhang, Yan, Wang, Wang, and Lian]{shen2025dast-difficultyadaptiveslowthinkinglarge}
Yi~Shen, Jian Zhang, Jieyun Huang, Shuming Shi, Wenjing Zhang, Jiangze Yan, Ning Wang, Kai Wang, and Shiguo Lian.
\newblock Dast: Difficulty-adaptive slow-thinking for large reasoning models, 2025.
\newblock URL \url{https://arxiv.org/abs/2503.04472}.

\bibitem[Hou et~al.(2025)Hou, Zhang, Ji, Liu, Qian, Andreas, and Chang]{hou2025thinkprune}
Bairu Hou, Yang Zhang, Jiabao Ji, Yujian Liu, Kaizhi Qian, Jacob Andreas, and Shiyu Chang.
\newblock Thinkprune: Pruning long chain-of-thought of llms via reinforcement learning.
\newblock \emph{arXiv preprint arXiv:2504.01296}, 2025.

\bibitem[Luo et~al.(2025{\natexlab{b}})Luo, Shen, He, Wang, Liu, Li, Tan, Cao, and Tao]{luo2025o1prunerlengthharmonizingfinetuningo1like}
Haotian Luo, Li~Shen, Haiying He, Yibo Wang, Shiwei Liu, Wei Li, Naiqiang Tan, Xiaochun Cao, and Dacheng Tao.
\newblock O1-pruner: Length-harmonizing fine-tuning for o1-like reasoning pruning, 2025{\natexlab{b}}.
\newblock URL \url{https://arxiv.org/abs/2501.12570}.

\bibitem[Team et~al.(2025)Team, Du, Gao, Xing, Jiang, Chen, Li, Xiao, Du, Liao, Tang, Wang, Zhang, Yuan, Lu, Tang, Sung, Wei, Lai, Guo, Zhu, Ding, Hu, Yang, Zhang, Yao, Zhao, Lu, Li, Yu, Gao, Zheng, Yuan, Chen, Guo, Su, Wang, Zhao, Zhang, Liu, Yan, Wu, Shi, Ye, Yu, Dong, Zhang, Ma, Pan, Gong, Liu, Ma, Wei, Cao, Huang, Jiang, Gao, Xiong, He, Huang, Xu, Wu, He, Wei, Jia, Wu, Xu, Zu, Zhou, Pan, Charles, Li, Hu, Liu, Chen, Wang, Liu, Qin, Liu, Yang, Bao, Du, Wu, Wang, Zhou, Wang, Li, Zhu, Zhang, Wang, Yang, Huang, Huang, Xu, Yang, and Lin]{kimi}
Kimi Team, Angang Du, Bofei Gao, Bowei Xing, Changjiu Jiang, Cheng Chen, Cheng Li, Chenjun Xiao, Chenzhuang Du, Chonghua Liao, Chuning Tang, Congcong Wang, Dehao Zhang, Enming Yuan, Enzhe Lu, Fengxiang Tang, Flood Sung, Guangda Wei, Guokun Lai, Haiqing Guo, Han Zhu, Hao Ding, Hao Hu, Hao Yang, Hao Zhang, Haotian Yao, Haotian Zhao, Haoyu Lu, Haoze Li, Haozhen Yu, Hongcheng Gao, Huabin Zheng, Huan Yuan, Jia Chen, Jianhang Guo, Jianlin Su, Jianzhou Wang, Jie Zhao, Jin Zhang, Jingyuan Liu, Junjie Yan, Junyan Wu, Lidong Shi, Ling Ye, Longhui Yu, Mengnan Dong, Neo Zhang, Ningchen Ma, Qiwei Pan, Qucheng Gong, Shaowei Liu, Shengling Ma, Shupeng Wei, Sihan Cao, Siying Huang, Tao Jiang, Weihao Gao, Weimin Xiong, Weiran He, Weixiao Huang, Weixin Xu, Wenhao Wu, Wenyang He, Xianghui Wei, Xianqing Jia, Xingzhe Wu, Xinran Xu, Xinxing Zu, Xinyu Zhou, Xuehai Pan, Y.~Charles, Yang Li, Yangyang Hu, Yangyang Liu, Yanru Chen, Yejie Wang, Yibo Liu, Yidao Qin, Yifeng Liu, Ying Yang, Yiping Bao, Yulun Du, Yuxin Wu, Yuzhi Wang, Zaida
  Zhou, Zhaoji Wang, Zhaowei Li, Zhen Zhu, Zheng Zhang, Zhexu Wang, Zhilin Yang, Zhiqi Huang, Zihao Huang, Ziyao Xu, Zonghan Yang, and Zongyu Lin.
\newblock Kimi k1.5: Scaling reinforcement learning with llms, 2025.
\newblock URL \url{https://arxiv.org/abs/2501.12599}.

\bibitem[Team(2025{\natexlab{a}})]{qwen3}
Qwen Team.
\newblock Qwen3, April 2025{\natexlab{a}}.
\newblock URL \url{https://qwenlm.github.io/blog/qwen3/}.

\bibitem[Team(2025{\natexlab{b}})]{qwq32b}
Qwen Team.
\newblock Qwq-32b: Embracing the power of reinforcement learning, March 2025{\natexlab{b}}.
\newblock URL \url{https://qwenlm.github.io/blog/qwq-32b/}.

\bibitem[Bercovich et~al.(2025)Bercovich, Levy, Golan, Dabbah, El-Yaniv, Puny, Galil, Moshe, Ronen, Nabwani, Shahaf, Tropp, Karpas, Zilberstein, Zeng, Singhal, Bukharin, Yian~Zhang, and Alexiuk]{bercovich2025llama-nemotron}
Akhiad Bercovich, Itay Levy, Izik Golan, Mohammad Dabbah, Ran El-Yaniv, Omri Puny, Ido Galil, Zach Moshe, Tomer Ronen, Najeeb Nabwani, Ido Shahaf, Oren Tropp, Ehud Karpas, Ran Zilberstein, Jiaqi Zeng, Soumye Singhal, Alexander Bukharin, ... Yian~Zhang, and Chris Alexiuk.
\newblock Llama-nemotron: Efficient reasoning models, 2025.
\newblock URL \url{https://arxiv.org/abs/2505.00949}.

\bibitem[Liu et~al.(2025)Liu, Chen, Li, Qi, Pang, Du, Lee, and Lin]{liu2025drgrpo}
Zichen Liu, Changyu Chen, Wenjun Li, Penghui Qi, Tianyu Pang, Chao Du, Wee~Sun Lee, and Min Lin.
\newblock Understanding r1-zero-like training: A critical perspective, 2025.
\newblock URL \url{https://arxiv.org/abs/2503.20783}.

\bibitem[Yu et~al.(2025)Yu, Zhang, Zhu, Yuan, Zuo, Yue, Fan, Liu, Liu, Liu, et~al.]{yu2025dapo}
Qiying Yu, Zheng Zhang, Ruofei Zhu, Yufeng Yuan, Xiaochen Zuo, Yu~Yue, Tiantian Fan, Gaohong Liu, Lingjun Liu, Xin Liu, et~al.
\newblock Dapo: An open-source llm reinforcement learning system at scale.
\newblock \emph{arXiv preprint arXiv:2503.14476}, 2025.

\bibitem[Rafailov et~al.(2023)Rafailov, Sharma, Mitchell, Manning, Ermon, and Finn]{rafailov2023direct}
Rafael Rafailov, Archit Sharma, Eric Mitchell, Christopher~D Manning, Stefano Ermon, and Chelsea Finn.
\newblock Direct preference optimization: Your language model is secretly a reward model.
\newblock \emph{Advances in Neural Information Processing Systems}, 36:\penalty0 53728--53741, 2023.

\bibitem[OpenAI(2024)]{chatgpt-o1}
OpenAI.
\newblock Chatgpt.
\newblock \url{https://openai.com/o1/}, 2024.

\bibitem[Google(2025{\natexlab{a}})]{gemini2.5pro}
Google.
\newblock Gemini 2.5 pro.
\newblock \url{https://cloud.google.com/vertex-ai/generative-ai/docs/models/gemini/2-5-pro}, 2025{\natexlab{a}}.

\bibitem[Abdin et~al.(2025)Abdin, Agarwal, Awadallah, Balachandran, Behl, Chen, de~Rosa, Gunasekar, Javaheripi, Joshi, Kauffmann, Lara, Mendes, Mitra, Nushi, Papailiopoulos, Saarikivi, Shah, Shrivastava, Vineet, Wu, Yousefi, and Zheng]{abdin2025phi4reasoning}
Marah Abdin, Sahaj Agarwal, Ahmed Awadallah, Vidhisha Balachandran, Harkirat Behl, Lingjiao Chen, Gustavo de~Rosa, Suriya Gunasekar, Mojan Javaheripi, Neel Joshi, Piero Kauffmann, Yash Lara, Caio César~Teodoro Mendes, Arindam Mitra, Besmira Nushi, Dimitris Papailiopoulos, Olli Saarikivi, Shital Shah, Vaishnavi Shrivastava, Vibhav Vineet, Yue Wu, Safoora Yousefi, and Guoqing Zheng.
\newblock Phi-4-reasoning technical report, 2025.
\newblock URL \url{https://arxiv.org/abs/2504.21318}.

\bibitem[Shao et~al.(2024)Shao, Wang, Zhu, Xu, Song, Bi, Zhang, Zhang, Li, Wu, and Guo]{shao2024deepseekmathpushinglimitsmathematical}
Zhihong Shao, Peiyi Wang, Qihao Zhu, Runxin Xu, Junxiao Song, Xiao Bi, Haowei Zhang, Mingchuan Zhang, Y.~K. Li, Y.~Wu, and Daya Guo.
\newblock Deepseekmath: Pushing the limits of mathematical reasoning in open language models, 2024.
\newblock URL \url{https://arxiv.org/abs/2402.03300}.

\bibitem[Liu and Zhang(2025)]{code-r1}
Jiawei Liu and Lingming Zhang.
\newblock Code-r1: Reproducing r1 for code with reliable rewards.
\newblock 2025.

\bibitem[Hu et~al.(2025)Hu, Liu, and Shen]{hu2025reinforceefficientrlhfalgorithm}
Jian Hu, Jason~Klein Liu, and Wei Shen.
\newblock Reinforce++: An efficient rlhf algorithm with robustness to both prompt and reward models, 2025.
\newblock URL \url{https://arxiv.org/abs/2501.03262}.

\bibitem[Ma et~al.(2025{\natexlab{a}})Ma, Wan, Yu, Fang, and Wang]{ma2025cotvalvelengthcompressiblechainofthoughttuning}
Xinyin Ma, Guangnian Wan, Runpeng Yu, Gongfan Fang, and Xinchao Wang.
\newblock Cot-valve: Length-compressible chain-of-thought tuning, 2025{\natexlab{a}}.
\newblock URL \url{https://arxiv.org/abs/2502.09601}.

\bibitem[Arora and Zanette(2025)]{arora2025efficientreasoning}
Daman Arora and Andrea Zanette.
\newblock Training language models to reason efficiently, 2025.
\newblock URL \url{https://arxiv.org/abs/2502.04463}.

\bibitem[Aytes et~al.(2025)Aytes, Baek, and Hwang]{aytes2025sketchofthoughtefficientllmreasoning}
Simon~A. Aytes, Jinheon Baek, and Sung~Ju Hwang.
\newblock Sketch-of-thought: Efficient llm reasoning with adaptive cognitive-inspired sketching, 2025.
\newblock URL \url{https://arxiv.org/abs/2503.05179}.

\bibitem[Han et~al.(2024)Han, Wang, Fang, Zhao, Ma, and Chen]{han2024token-budget}
Tingxu Han, Zhenting Wang, Chunrong Fang, Shiyu Zhao, Shiqing Ma, and Zhenyu Chen.
\newblock Token-budget-aware llm reasoning.
\newblock \emph{arXiv preprint arXiv:2412.18547}, 2024.

\bibitem[Ma et~al.(2025{\natexlab{b}})Ma, He, Snell, Griggs, Min, and Zaharia]{ma2025nothink}
Wenjie Ma, Jingxuan He, Charlie Snell, Tyler Griggs, Sewon Min, and Matei Zaharia.
\newblock Reasoning models can be effective without thinking, 2025{\natexlab{b}}.
\newblock URL \url{https://arxiv.org/abs/2504.09858}.

\bibitem[Team(2024)]{qwen2.5}
Qwen Team.
\newblock Qwen2.5: A party of foundation models, September 2024.
\newblock URL \url{https://qwenlm.github.io/blog/qwen2.5/}.

\bibitem[Google(2025{\natexlab{b}})]{google2024gemini2.5flash}
Google.
\newblock Gemini 2.5 flash.
\newblock \url{https://developers.googleblog.com/en/start-building-with-gemini-25-flash/}, 2025{\natexlab{b}}.

\bibitem[Yu et~al.(2024)Yu, Xu, Weston, and Kulikov]{yu2024distilling21}
Ping Yu, Jing Xu, Jason Weston, and Ilia Kulikov.
\newblock Distilling system 2 into system 1, 2024.
\newblock URL \url{https://arxiv.org/abs/2407.06023}.

\bibitem[{International Conference on Artificial Intelligence in Medicine}()]{aime25}
{International Conference on Artificial Intelligence in Medicine}.
\newblock The 23rd international conference on artificial intelligence in medicine (aime 2025).
\newblock \url{https://aime25.aimedicine.info/}.
\newblock Accessed: 2025-06-10.

\bibitem[Lightman et~al.(2023)Lightman, Kosaraju, Burda, Edwards, Baker, Lee, Leike, Schulman, Sutskever, and Cobbe]{lightman2023let}
Hunter Lightman, Vineet Kosaraju, Yuri Burda, Harrison Edwards, Bowen Baker, Teddy Lee, Jan Leike, John Schulman, Ilya Sutskever, and Karl Cobbe.
\newblock Let's verify step by step.
\newblock In \emph{The Twelfth International Conference on Learning Representations}, 2023.

\bibitem[Cobbe et~al.(2021)Cobbe, Kosaraju, Bavarian, Chen, Jun, Kaiser, Plappert, Tworek, Hilton, Nakano, Hesse, and Schulman]{cobbe2021gsm8k}
Karl Cobbe, Vineet Kosaraju, Mohammad Bavarian, Mark Chen, Heewoo Jun, Lukasz Kaiser, Matthias Plappert, Jerry Tworek, Jacob Hilton, Reiichiro Nakano, Christopher Hesse, and John Schulman.
\newblock Training verifiers to solve math word problems.
\newblock \emph{arXiv preprint arXiv:2110.14168}, 2021.

\bibitem[{Mathematical Association of America}()]{maa2025amc}
{Mathematical Association of America}.
\newblock American mathematics competitions (amc).
\newblock \url{https://maa-amc.org/student-programs/amc/}.
\newblock Accessed: 2025-06-10.

\bibitem[Rein et~al.(2024)Rein, Hou, Stickland, Petty, Pang, Dirani, Michael, and Bowman]{rein2024gpqa}
David Rein, Betty~Li Hou, Asa~Cooper Stickland, Jackson Petty, Richard~Yuanzhe Pang, Julien Dirani, Julian Michael, and Samuel~R Bowman.
\newblock Gpqa: A graduate-level google-proof q\&a benchmark.
\newblock In \emph{First Conference on Language Modeling}, 2024.

\bibitem[Jain et~al.(2024)Jain, Han, Gu, Li, Yan, Zhang, Wang, Solar-Lezama, Sen, and Stoica]{jain2024livecodebench}
Naman Jain, King Han, Alex Gu, Wen-Ding Li, Fanjia Yan, Tianjun Zhang, Sida Wang, Armando Solar-Lezama, Koushik Sen, and Ion Stoica.
\newblock Livecodebench: Holistic and contamination free evaluation of large language models for code.
\newblock \emph{arXiv preprint arXiv:2403.07974}, 2024.
\newblock \doi{10.48550/arXiv.2403.07974}.

\bibitem[von Werra et~al.(2020)von Werra, Belkada, Tunstall, Beeching, Thrush, Lambert, Huang, Rasul, and Gallouédec]{vonwerra2022trl}
Leandro von Werra, Younes Belkada, Lewis Tunstall, Edward Beeching, Tristan Thrush, Nathan Lambert, Shengyi Huang, Kashif Rasul, and Quentin Gallouédec.
\newblock Trl: Transformer reinforcement learning.
\newblock \url{https://github.com/huggingface/trl}, 2020.

\end{thebibliography}
}
\onecolumn
\appendix

\section{Related Work}

\paragraph{Reinforcement Learning For Reasoning.}

Reinforcement learning (RL) has emerged as a pivotal technique in the post-training phase of Large Language Models (LLMs), demonstrating significant potential in enhancing their reasoning abilities. A landmark work in this domain is OpenAI's o1 model \citep{chatgpt-o1}. As the first large-scale application of RL for reasoning, o1 achieved state-of-the-art reasoning capabilities at the time of its release. Shortly after, Deepseek-R1 \citep{deepseekai2025deepseek-r1} was introduced as the first open-source model to match the performance of o1, significantly advancing the development and popularization of RL-based reasoning techniques. This technical approach has led to the emergence of numerous powerful Large Reasoning Models (LRMs), such as Gemini 2.5 \citep{gemini2.5pro}, QwQ \citep{qwq32b}, and Phi-4 \citep{abdin2025phi4reasoning}. Recently, Reinforcement Learning with Verifiable Rewards (RLVR) has been shown to be an effective method for significantly improving model reasoning abilities in domains like mathematics and programming \citep{shao2024deepseekmathpushinglimitsmathematical, code-r1}.  Concurrently, the research community has proposed various advanced RL algorithms to optimize the post-training process, including GRPO \citep{shao2024deepseekmathpushinglimitsmathematical}, Reinforce++ \citep{hu2025reinforceefficientrlhfalgorithm}, DAPO \citep{yu2025dapo}, and Dr.GRPO \citep{liu2025drgrpo}. These algorithmic innovations have continuously improved the efficiency and effectiveness of the RL training process. Our proposed method, \method, builds upon these foundational works with specific adjustments aimed at pruning redundant sequence in the reasoning process, achieving a more favorable trade-off between efficacy and efficiency.

\paragraph{Efficient Reasoning.}

While elaborate reasoning is more likely to yield a correct answer, its verbose thought process significantly increases time and computational costs, a phenomenon termed \textit{overthinking} \citep{chen2024overthinking}. To mitigate this issue, researchers have proposed various solutions from different perspectives \citep{sui2025stop-overthinking-survey}. These approaches can be broadly categorized. The first category involves directly constraining the redundancy of the reasoning process through length control, with typical examples like CoT-Valve \citep{ma2025cotvalvelengthcompressiblechainofthoughttuning} and L1 \citep{aggarwal2025l1controllinglongreasoning}. A second category of methods focuses on enabling the model to adapt its reasoning depth according to the difficulty of the query. For instance, Adar1 \citep{luo2025adar1} and DAST \citep{shen2025dast-difficultyadaptiveslowthinkinglarge} construct preference datasets to train the model to generate reasoning sequences that match the problem's complexity. A third category integrates efficiency considerations into the reinforcement learning framework. Works like O1-Pruner \citep{luo2025o1prunerlengthharmonizingfinetuningo1like}, ThinkPrune \citep{hou2025thinkprune}, Training \citep{arora2025efficientreasoning}, and Kimi \citep{kimi} incorporate length-based penalties into the reward function, incentivizing the model to produce more concise reasoning while maintaining accuracy. Furthermore, there are also training-free methods that enhance reasoning efficiency on-the-fly through sophisticated prompting techniques, such as Sketch-of-Thought \citep{aytes2025sketchofthoughtefficientllmreasoning}, Token-Budget \citep{han2024token-budget}, and No Think \citep{ma2025nothink}.
\section{Details of LC-Extractor}
\label{Appendix: LC-Extractor}

\begin{figure*}[!t]
    \centering
    \includegraphics[width=\textwidth]{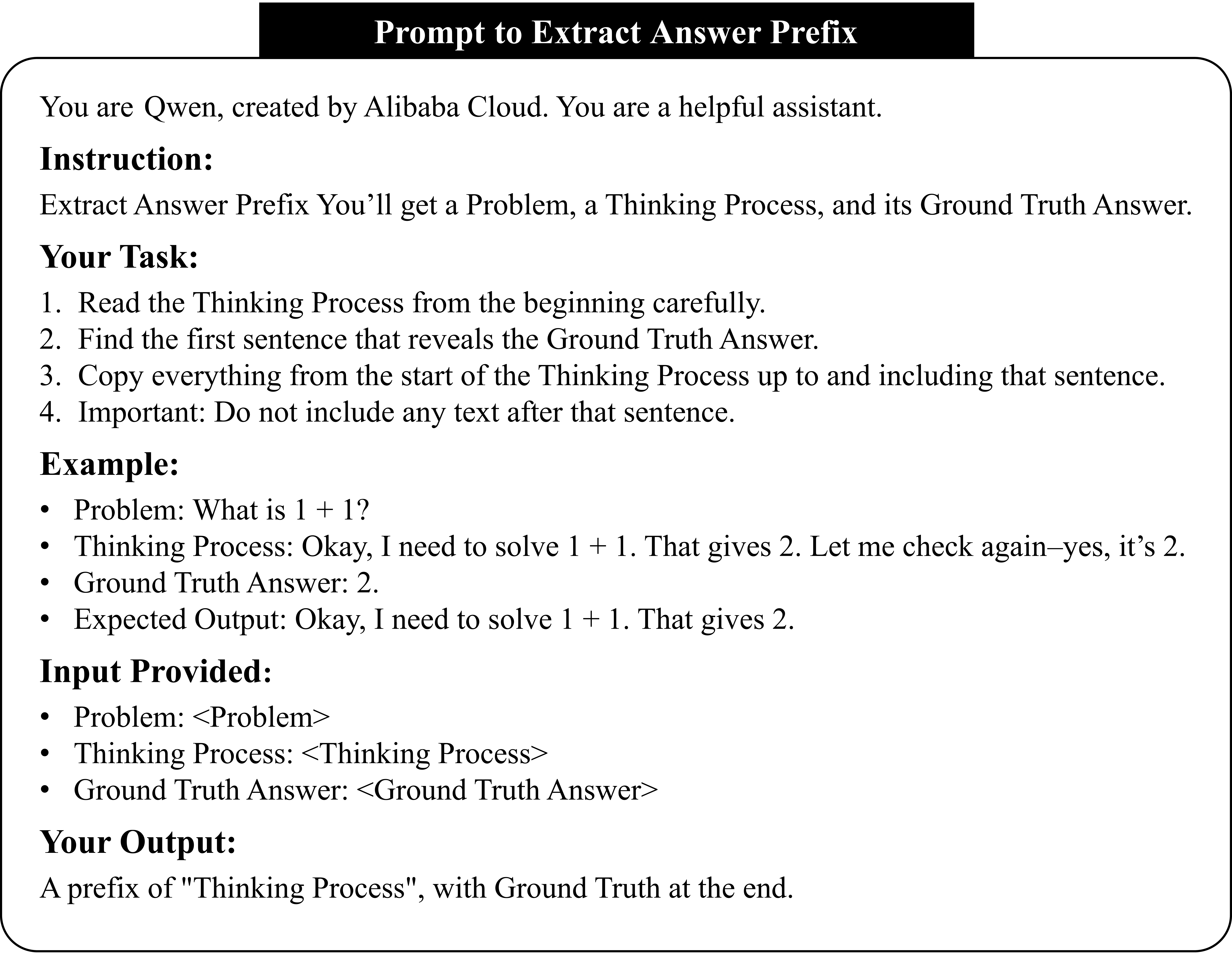}
    \caption{Our prompt for extraction of answer prefix.}
    \label{fig:prompt}
\end{figure*}

\begin{figure*}[!t]
    \centering
    \includegraphics[width=\textwidth]{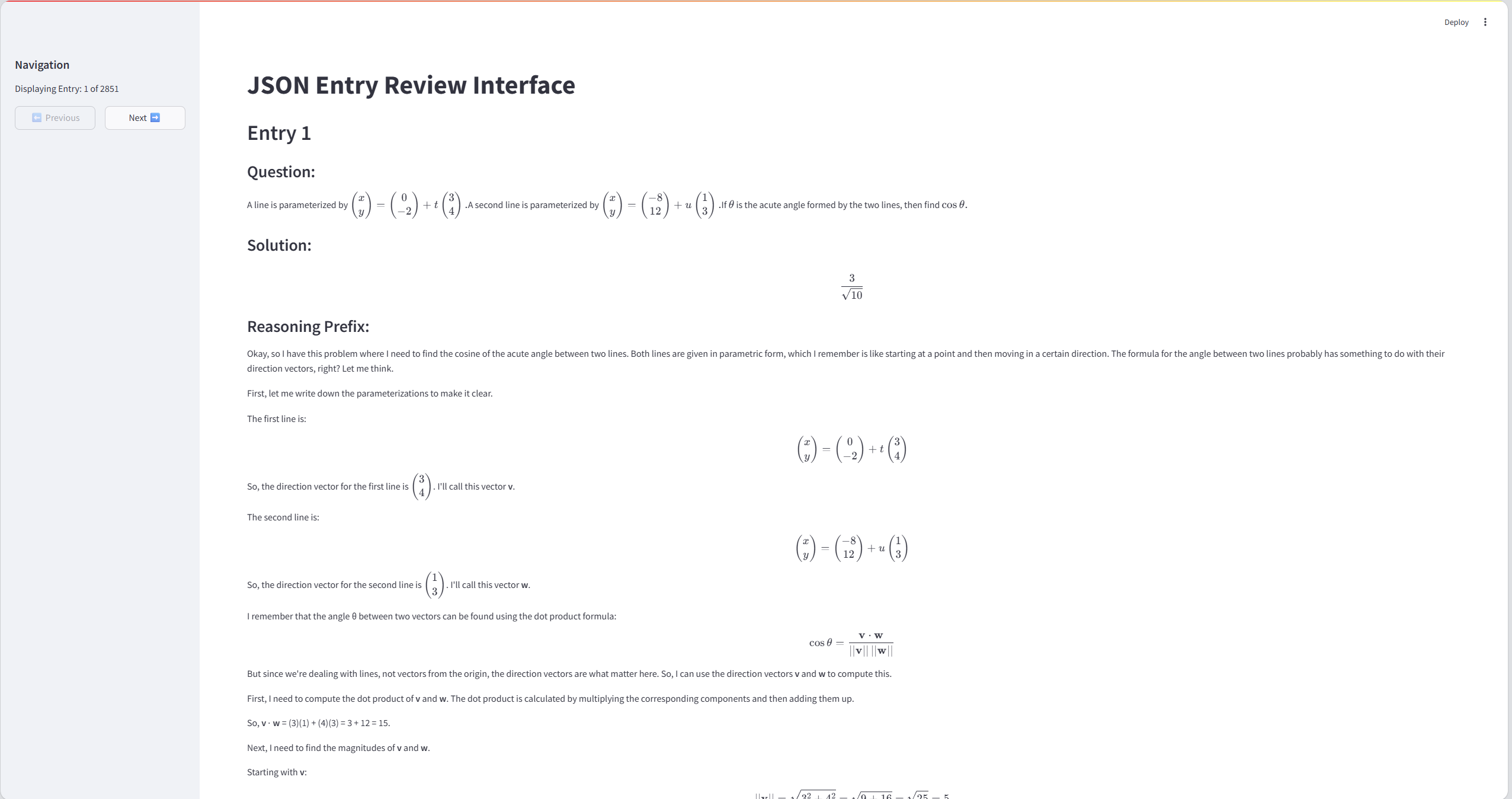}
    \caption{The annotation tool to evaluate the LC-Extratcor.}
    \label{fig:dabiao}
    
\end{figure*}

We train Qwen-2.5-3B-Instruct \citep{qwen2.5} as the LC-Extractor model. We construct a dataset consisting of 5,000 $<$\textit{Question, Thinking Process, Answer}$>$ triplets from MATH dataset and identify the position of the first correct token using Gemini-2.5-Flash \citep{google2024gemini2.5flash}, followed by rigorous rule-based filtering.  We then distill this knowledge into a smaller model through training for 2 epochs with these curated samples. \extractor's effectiveness is validated on a 100-sample test set, achieving 98\% accuracy as confirmed by human evaluation as shown in in Figure \ref{fig:dabiao}. LC-Extractor model is activated by the prompt in Figure \ref{fig:prompt}. 

\section{Detailed Experiment Setups}
\label{Appendix: Experiment Setups}
\subsection{Model}

We use \textbf{DeepSeek-R1}\citep{deepseekai2025deepseek-r1}, \textbf{Qwen3-32B}\citep{qwen3}, \textbf{QwQ-32B}\citep{qwq32b}, \textbf{Llama-3.3-Nemotrom-Super-49B-V1}\citep{bercovich2025llama-nemotron}, \textbf{Distill-Qwen-7B}, \textbf{Distill-Qwen-1.5B}\citep{yu2024distilling21}, and \textbf{Qwen-2.5-3B-Instruct}\citep{qwen2.5} models in our paper. We introduce their licenses and key characteristics as follows:

\begin{itemize}[leftmargin=*, itemsep=0pt]
\item \textbf{DeepSeek-R1.} An open-source 671 B→37 B MoE reasoning model trained largely through reinforcement learning, which elicits self-verification, reflection and lengthy chain-of-thought traces while supporting 128K-token context; it matches proprietary o1 on math / code benchmarks using only public data.
\item \textbf{Qwen3-32B.} The 32.8 B‐parameter third-generation Qwen model that toggles between “thinking” and “non-thinking” modes, delivering state-of-the-art reasoning, multilingual chat and up to 131 K context in a single dense checkpoint.
\item \textbf{QwQ-32B.} A medium-sized Qwen reasoning variant refined with SFT + RL; provides explicit <think> traces, 131 K context and DeepSeek-R1–level accuracy on hard evaluations.
\item \textbf{Llama-3.3-Nemotrom-Super-49B-V1.} NVIDIA’s NAS-pruned 49 B derivative of Llama-3.3-70B, post-trained for reasoning, RAG and tool calling; couples 128 K context with single-H100 deployment efficiency for cost-sensitive production.
\item \textbf{Deepseek-R1-Distill-Qwen-7B.} A 7 B dense checkpoint distilled from DeepSeek-R1 onto the Qwen2.5 backbone, pushing small-model MATH-500 pass\@1 beyond 92 \% and surpassing o1-mini on several reasoning suites while remaining laptop-friendly.
\item \textbf{Deepseek-R1-Distill-Qwen-1.5B.} An ultra-compact 1.5 B model distilled from R1 that preserves chain-of-thought and achieves 83.9 \% pass\@1 on MATH-500, bringing competitive analytical power to edge and mobile deployments.
\item \textbf{Qwen-2.5-3B-Instruct.} A 3.09 B instruction-tuned model with 128 K context, strengthened coding/math skills and multilingual support, designed as a lightweight yet controllable chat foundation for downstream tasks.
\end{itemize}

\subsection{Dataset}
We benchmark on the \textbf{AIME25} \citep{aime25}, \textbf{MATH500} \citep{lightman2023let}, \textbf{GSM8K} \citep{cobbe2021gsm8k}, \textbf{OlympiadBench} \citep{sun2025olym-math}, \textbf{AMC} \citep{maa2025amc}, \textbf{GPQA Diamond} \citep{rein2024gpqa} and \textbf{LiveCodeBench} \citep{jain2024livecodebench} benchmarks in our paper. We introduce them as follows:

\begin{itemize}[leftmargin=*, itemsep=0pt]
    \item \textbf{AIME25.} A benchmark with 30 questions distilled from twenty-five years of \emph{American Invitational Mathematics Examination} papers.  Each item is a three-digit short-answer problem that probes upper-secondary algebra, geometry, combinatorics.

    \item \textbf{MATH500.} A 500-problem evaluation slice covering the full subject breadth of the original \emph{MATH} competition corpus.  Balanced across difficulty tiers and topics, it serves as a rigorous yardstick for advanced high-school and early undergraduate mathematical reasoning, without the runtime burden of the complete 12k-question set.

    \item \textbf{GSM8K.} The widely-adopted \emph{Grade-School Math 8K} benchmark of 1,319 everyday word-problems.  Requiring multi-step arithmetic and commonsense, GSM8K remains the de-facto standard for assessing chain-of-thought quality on conversational math tasks.

    \item \textbf{Olympiad.} A curated collection of roughly 3 k national and international mathematics-olympiad problems.  Predominantly proof-style or numeric-answer challenges, this benchmark gauges creative, non-routine reasoning at the highest pre-university level.

    \item \textbf{AMC.} An aggregate of 83 from the \emph{American Mathematics Competitions 10/12}.  Spanning 2000–2024, it offers a longitudinal benchmark on foundational secondary-school math.

    \item \textbf{GPQA Diamond}. A benchmark with 198 graduate-level Google-proof multiple-choice questions requiring deep domain expertise and multi-step reasoning, curated by researchers from New York University, CohereAI, and Anthropic; evaluated in closed-book and open-book settings using accuracy as the metric.
    
    \item \textbf{LiveCodeBench}. A dynamic, contamination-free coding benchmark originally hosting 511 problems (release v2) collected from LeetCode, AtCoder, and CodeForces, designed by UC Berkeley, MIT, and Cornell researchers to holistically assess LLMs' code generation, execution, and test prediction capabilities using Pass@K.

\end{itemize}

\subsection{settings}

We used a mixed-difficulty dataset, combining past AIME competition problems with the
MATH dataset in an approximate 1:2 ratio to create 2500 training data. We use Trl\citep{vonwerra2022trl} framework to train models. Both models are trained with 4 * A800-80G GPUs and the hyperparameters are presented in Table \ref{tab:hyper}.

\begin{table}[h]
\centering
\caption{Hyperparameters for \method training.}
\label{tab:hyper}
\begin{tabular}{lcc}
\hline
\textbf{Hyperparameter} & \textbf{R1-Distill-Qwen-7B} & \textbf{R1-Distill-Qwen-1.5B} \\
\hline
cutoff\_len & 8192 & 8192 \\
batch\_size & 32 & 32 \\
learning\_rate & 3.0e-6 & 2.0e-6 \\
num\_train\_epochs & 1.0 & 1.0 \\
$\alpha$ & 1.0 & 1.0 \\
$\beta$ & 0.04 & 0.04 \\
$\gamma$ & 1.0 & 1.0 \\
num\_generations & 6 & 8 \\
$\epsilon$ & 0.2 & 0.2 \\
\hline
\end{tabular}
\end{table}

\paragraph{Baseline settings.} We compare \method with 5 baseline---SFT, DPO, O1-Pruner, ThinkPrune, SFT+O1-Pruner. The last hybrid method shares same settings with each method, so we give out the settings of first four methods.
\begin{itemize}
    \item SFT. We construct training dataset by extracting the valid thinking process to reconstruct a concise version of sequences sampled by themselves on MATH dataset. We set cutoff\_len=8192, epoch=1, learning\_rate = 3.0e-6, max\_samples = 5000.
    \item DPO. We construct preference training dataset by sampling 8 times on MATH dataset and choose the longest sample to be negative and shortest sample to be positive. We set cutoff\_len=8192, epoch=2, learning\_rate = 5e-6, max\_samples = 5000.
    \item O1-Pruner. We use the given python scripts to construct weight training dataset, with cutoff\_len=4096, epoch=2, learning\_rate = 2.0e-7, max\_samples = 10000.
    \item ThinkPrune-3K. We reproduce the training process on ThinkPrune-length3000 dataset, with size 2470. We set cutoff\_len=8192, epoch=2, learning\_rate = 2.0e-6, num\_generations=8, batch\_size=32.
\end{itemize}
\section{Case Study}
\label{Appendix: Case Study}

We make some case studies to compare \method with O1-Pruner\citep{luo2025o1prunerlengthharmonizingfinetuningo1like} method and the base model. These case studies are shown in Figure \ref{fig:case1} and Figure \ref{fig:case2}.

\begin{figure*}[!t]
    \centering
    \includegraphics[width=0.9\textwidth]{Figures/c1.pdf}
    \caption{Case study of the comparison of \method and O1-Pruner.}
    \label{fig:case1}
\end{figure*}

\begin{figure*}[!t]
    \centering
    \includegraphics[width=0.9\textwidth]{Figures/c2.pdf}
    \caption{Case study of the comparison of \method and the original model.}
    \label{fig:case2}
\end{figure*}

\end{document}